# An improved axiomatic definition of information granulation

Ping Zhu[a]

[a]*School of Science, Beijing University of Posts and Telecommunications, Beijing 100876, China*

**Abstract**

To capture the uncertainty of information or knowledge in information systems, various information granulations, also known as knowledge granulations, have been proposed. Recently, several axiomatic definitions of information granulation have been introduced. In this paper, we try to improve these axiomatic definitions and give a universal construction of information granulation by relating information granulations with a class of functions of multiple variables. We show that the improved axiomatic definition has some concrete information granulations in the literature as instances.

## 1. Introduction

Information systems [8, 1], also called knowledge representation systems, are a formalism for representing knowledge about some objects in terms of attributes (e.g., color) and values of attributes (e.g., green). Any information system can be represented by a data table containing rows labeled by objects of interest, columns labeled by attributes, and entries of the table indicating attribute values. An arbitrary set of attributes determines a classification of objects. Each subclass under a classification is called an information (or knowledge) granule. In data analysis, a basic problem we are interested in is to reason about the accessible granules of information. To this end, various information (or knowledge) granulations, as an average measure of information granules, have been proposed in [2, 6, 7, 9, 10].

Since the concrete information granulations in [2, 6, 7, 9, 10] share some common properties, Liang et al. [3, 4, 5] and Zhao et al. [11] attempted to unify the definitions by some axiomatic approaches. Although these axiomatic definitions of information granulations in [3, 4, 5, 11] characterize efficiently some important aspects of information granulations, they have slight imperfections in the formalization and mathematical rigor, which will be explained and discussed thoroughly in the body of the paper. The aim of this paper is thus to provide an improved, stronger version of these axiomatic definitions of information granulation. By making a connection between information granulations and a class of functions of multiple variables, we introduce a universal construction of information granulation. Based on the construction, we show that various information granulations in the literature [2, 6, 7, 9, 10] are some instances of our axiomatic definition.

The structure of the paper is as follows. After briefly recalling some notions and notations of information systems in Section 2, we review the existent axiomatic definitions of information granulation in Section 3. Section 4 explores an improvement of the axiomatic definitions, consisting of a new axiomatic definition and a case study. We provide a universal construction of information granulation in Section 5 and conclude the paper in Section 6.

## 2. Information systems

In this section, we briefly recall the definitions of complete and incomplete information systems. For a detailed introduction to the relevant notions, the reader may refer to [8, 1, 5].

An information system is a pair $S = (U, A)$ of non-empty, finite sets $U$ and $A$, where $U$ is the universe of objects and $A$ is the set of attributes. An attribute is a mapping $a : U \longrightarrow V_a$, where $V_a$ is the set of values of attribute $a$, called the domain of $a$.

An information system $S = (U, A)$ is said to be complete if for any $a \in A$, every element of $V_a$ is a definite value; otherwise, it is called incomplete. In other words, we say that $S = (U, A)$ is an incomplete information system if there exists an attribute $a \in A$ such that $V_a$ contains a null value. As usual, we write $*$ for the null value.

For a complete information system $S = (U, A)$, any subset $P$ of $A$ determines a binary relation $Ind(P)$ on $U$, called an indiscernibility relation, defined by

$$Ind(P) = \{(x, y) \in U \times U \mid \forall\, a \in P, a(x) = a(y)\}.$$

Clearly, $Ind(P) = \bigcap_{a \in P} Ind(\{a\})$, and moreover, $Ind(P)$ is an equivalence relation. Each equivalence class of $Ind(P)$ is called an information granule. The family of all equivalence classes of $Ind(P)$, i.e., the partition determined by $Ind(P)$, is denoted by $U/Ind(P)$. In particular, if $U/Ind(P) = \{\{x\} \mid x \in U\}$, it is called an identity relation, denote by $\omega$; if $U/Ind(P) = \{U\}$, it is called a universal relation, denote by $\delta$.

For an incomplete information system $S = (U, A)$, any subset $P$ of $A$ also determines a binary relation $Sim(P)$ on $U$, called similarity relation or tolerance relation, defined by

$$Sim(P) = \{(x, y) \in U \times U \mid \forall\, a \in P, a(x) = a(y)$$
$$\text{or } a(x) = * \text{ or } a(y) = *\}.$$

Obviously, $Sim(P) = \bigcap_{a \in P} Sim(\{a\})$, but it is not necessarily an equivalence relation. It satisfies the reflexive and symmetric conditions, but in general the transitivity does not hold. Denote by $S_P(x)$ the set $\{y \in U \mid (x, y) \in Sim(P)\}$ and by $\tilde{P}$ the set $\{S_P(x) \mid x \in U\}$. Then $\tilde{P}$ yields a cover of $U$, that is, $U = \bigcup_{x \in U} S_P(x)$ and $S_P(x) \neq \emptyset$ for any $x \in U$. In particular, if $\tilde{P} = \omega = \{\{x\} \mid x \in U\}$, it is called an identity relation; if $U/Ind(P) = \delta = \{U\}$, it is called a universal relation. Every tolerance class $S_P(x)$ is also called an information granule. For any $P \subseteq A$, we write $\alpha(P)$ for the vector $(S_P(x_1), S_P(x_2), \ldots, S_P(x_{|U|}))$, where $|U|$ stands for the cardinality of $U$. Thus, $\alpha$ is a mapping from $\mathscr{P}(A)$ to $\mathscr{P}(U)^{|U|}$, where $\mathscr{P}(A)$ is the power set of $A$ and $\mathscr{P}(U)^{|U|}$ is the cartesian product of $|U|$ components.

For later need, let us unify a notation. Let $S = (U, A)$ be a complete information system and $P \subseteq A$ with $U/Ind(P) = \{X_1, X_2, \ldots, X_m\}$. If we view $S = (U, A)$ as an incomplete information system without null values, then for every $x \in U$, we get a tolerance class $S_P(x)$. As a result, it gives rise to a vector $\alpha(P) = (S_P(x_1), S_P(x_2), \ldots, S_P(x_{|U|}))$. It turns out that if $X_i = \{x_{i1}, x_{i2}, \ldots, x_{is_i}\}$ ($i = 1, 2, \ldots, m$), then

$$X_i = S_P(x_{i1}) = S_P(x_{i2}) = \cdots = S_P(x_{is_i}), \tag{1}$$

$$\sum_{j=1}^{s_i} |S_P(x_{ij})| = |X_i|^2 = s_i^2, \tag{2}$$

$$\sum_{i=1}^{m} s_i = |U|. \tag{3}$$

In particular, the identity and universal relations give the vectors $(\{x_1\}, \{x_2\}, \ldots, \{x_{|U|}\})$ and $(U, U, \ldots, U)$, respectively. Summarily, whether $S = (U, A)$ is complete or not, to every $P \subseteq A$ we can always associate a vector $\alpha(P)$.

## 3. Existing axiomatic definitions of information granulation

In this section, we review the four axiomatic definitions of information granulation in the literature. Three of these definitions are due to Liang and Quan [3, 4, 5], and the remaining one is due to Zhao, Yang, and Gao [11]. We remark that the terms of information granulation and knowledge granulation used in [3, 4, 5, 11] just say the same thing.

*3.1. The first axiomatic definition*

Let us begin with some notations in [3].

Assume that $S = (U, A)$ is an information system. For any $P \subseteq A$, write $K(P)$ for the multiset[0] $\{S_P(x_1), S_P(x_2), \ldots, S_P(x_{|U|})\}$. A partial preorder $\preceq''$ on $\mathscr{P}(A)$ is defined as follows: For any $P, Q \in \mathscr{P}(A)$, $P \preceq'' Q$ if and only if there exists a sequence of elements of $K(Q)$, say, $S_Q(x'_1), S_Q(x'_2), \ldots, S_Q(x'_{|U|})$, such that $|S_P(x_i)| \leq |S_Q(x'_i)|$ for all $i \in \{1, 2, \ldots, |U|\}$. If there is a sequence $S_Q(x'_1), S_Q(x'_2), \ldots, S_Q(x'_{|U|})$ of elements of $K(Q)$ such that $|S_P(x_i)| < |S_Q(x'_i)|$ for every $i \in \{1, 2, \ldots, |U|\}$, then we denote it by $P \prec'' Q$. Note that $\preceq''$ is only a partial preorder, that is, it is reflexive and transitive, but not necessarily antisymmetric.

To the best of our knowledge, the following is the first axiomatic definition of information granulation.

**Definition 1** ([3]). *Let $S = (U, A)$ be an information system and $G$ a function from $\mathscr{P}(A)$ to $\mathbb{R}$, the set of real numbers. We say that $G$ is an information granulation of $S$ if $G$ satisfies the following conditions:*

*(1) $G(P) \geq 0$ for any $P \in \mathscr{P}(A)$ (Nonnegativity).*
*(2) $G(P) = G(Q)$ for any $P, Q \in \mathscr{P}(A)$ if there exists a bijective mapping $f : K(P) \longrightarrow K(Q)$ such that $|S_P(x_i)| = |f(S_P(x_i))|$ for all $i \in \{1, 2, \ldots, |U|\}$ (Invariability).*
*(3) $G(P) < G(Q)$ for any $P, Q \in \mathscr{P}(A)$ with $P \prec'' Q$ (Monotonicity).*

One of the advantages of this definition is that it takes into account the cardinality of tolerance classes, which is an important factor in all the existing concrete definitions of information granulation in [2, 6, 7, 9, 10]. The weakness is that the order $\prec''$ is very strong. In fact, when defining $P \prec'' Q$, we do not need that $|S_P(x_i)| < |S_Q(x'_i)|$ for every $i \in \{1, 2, \ldots, |U|\}$; one inequality is enough. In addition, the mapping between multisets, which seems infrequent, remains yet to be defined explicitly.

*3.2. The second axiomatic definition*

In [4], Liang and Qian proposed another axiomatic definition of information granulation. Keeping the notation in Section 3.1, we can state the definition as follows.

**Definition 2** ([4]). *Let $S = (U, A)$ be an information system and $G$ a function from $\mathscr{P}(A)$ to $\mathbb{R}$. We say that $G$ is an information granulation of $S$ if $G$ satisfies the following conditions:*

*(1) $G(P) \geq 0$ for any $P \in \mathscr{P}(A)$ (Nonnegativity).*
*(2) $G(P) = G(Q)$ for any $P, Q \in \mathscr{P}(A)$ if there exists a bijective mapping $f : K(P) \longrightarrow K(Q)$ such that $|S_P(x_i)| = |f(S_P(x_i))|$ for all $i \in \{1, 2, \ldots, |U|\}$ (Invariability).*
*(3) $G(P) \leq G(Q)$ for any $P, Q \in \mathscr{P}(A)$ with $P \preceq'' Q$ (Monotonicity).*

Compared with Definition 1, the unique difference appears in the monotonicity: the orders $\prec''$ and $<$ are relaxed to $\preceq''$ and $\leq$, respectively. This relaxation leads to that the information granulation $G$ may take constant values, which is not desirable.

*3.3. The third axiomatic definition*

More recently, Liang and Qian provided a new axiomatic definition of information granulation in [5].

In order to state the definition, we need to introduce an order at first.

Let $S = (U, A)$ be an information system. Recall that for any $P \subseteq A$, $\alpha(P)$ denotes the vector $(S_P(x_1), S_P(x_2), \ldots, S_P(x_{|U|}))$. Let us write $Im(\alpha)$ for the image of $\alpha$, namely, $Im(\alpha) = \{\alpha(P) \mid P \subseteq A\}$. A partial order $\preceq'$ on $Im(\alpha)$ is defined as follows [5]: For any $\alpha(P), \alpha(Q) \in Im(\alpha)$, $\alpha(P) \preceq' \alpha(Q)$ if and only if $S_P(x_i) \subseteq S_Q(x_i)$ for all $i \in \{1, 2, \ldots, |U|\}$. If both $\alpha(P) \preceq' \alpha(Q)$ and $\alpha(Q) \preceq' \alpha(P)$ hold, then we get that $\alpha(P) = \alpha(Q)$. If $\alpha(P) \preceq' \alpha(Q)$ and $\alpha(P) \neq \alpha(Q)$, then it is denoted by $\alpha(P) \prec' \alpha(Q)$.

**Definition 3** ([5]). *Let $S = (U, A)$ be an information system and $G$ a function from $\mathscr{P}(A)$ to $\mathbb{R}$. We say that $G$ is an information granulation of $S$ if $G$ satisfies the following conditions:*

*(1) $G(P) \geq 0$ for any $P \in \mathscr{P}(A)$ (Nonnegativity).*
*(2) $G(P) = G(Q)$ for any $P, Q \in \mathscr{P}(A)$ satisfying $\alpha(P) = \alpha(Q)$ (Invariability).*
*(3) $G(P) < G(Q)$ for any $P, Q \in \mathscr{P}(A)$ with $\alpha(P) \prec' \alpha(Q)$ (Monotonicity).*

Now, the monotonicity is handled well, but as we will see, the partial order is somewhat strong. The cardinality of tolerance classes, a significant ingredient of the concrete definitions of information granulation, is missing from such an axiomatic definition.

---

[0] It follows from the context of [3] that $K(P)$ should be a multiset, although it was not stressed there.

*3.4. The fourth axiomatic definition*

In [11], Zhao et al. criticized Definitions 1 and 2 for allowing information granulation to be constant functions and not being bounded by 0 and 1. Indeed, allowing information granulation to be constant is a defect, but we do not believe that allowing information granulation not to be bounded by 0 and 1 is really a matter, which will be demonstrated in the next section.

Let us introduce the axiomatic definition of information granulation in [11], which is based on a complete information system.

**Definition 4** ([11]). *Let $S = (U, A)$ be a complete information system and $G$ a function from $\mathscr{P}(A)$ to $\mathbb{R}$. We say that $G$ is an information granulation of $S$ if $G$ satisfies the following conditions:*

(1) *$G(P) = G(Q)$ for any $P, Q \in \mathscr{P}(A)$ if there exists a bijective mapping $f : U/Ind(P) \longrightarrow U/Ind(Q)$ such that $|X| = |f(X)|$ for any $X \in U/Ind(P)$ (Invariability).*
(2) *$G(P) < G(Q)$ for any $P, Q \in \mathscr{P}(A)$ with $P \prec Q$ (Monotonicity).*
(3) *For any $P \in \mathscr{P}(A)$, if $\tilde{P} = \omega = \{\{x\} \mid x \in U\}$, then $G(P) = 0$; if $U/Ind(P) = \delta = \{U\}$, then $G(P) = 1$ (Boundedness).*

In the above definition, the order $\prec$ has not been stated in [11]. In light of the context there, it should follow from the standard ordering on partitions. That is, $P \prec Q$ if and only if $P \preceq Q$ and $U/Ind(P) \neq U/Ind(Q)$, where $P \preceq Q$ if and only if each set in $U/Ind(P)$ is a subset of some set in $U/Ind(Q)$.

Beyond all question, Definition 4 is most appropriate for the axiomatic definition of information granulation in complete information systems. Further, the authors of [11] remarked that this definition is also established in incomplete information systems. However, this remark is not convincing. Observe that if $S = (U, A)$ is incomplete, then the partitions $U/Ind(P)$ and $U/Ind(Q)$ degenerate into covers. Unfortunately, it does not seem direct to apply the bijection $f$ and the order $\prec$ for covers. For example, in the proofs of some primary results such as Theorem 3.1 in [11], the authors used the fact that $P \prec Q$ implies $|U/Ind(P)| > |U/Ind(Q)|$. Such a fact does not hold in general when $U/Ind(P)$ and $U/Ind(Q)$ are covers.

## 4. An improved axiomatic definition of information granulation

This section is devoted to improving the existing axiomatic definitions of information granulation. We first give a new definition of information granulation and compare it with the existing ones reviewed in the last section. Then, as an example, we show that two concrete information granulations in [2, 6] fulfil our definition.

*4.1. Definition*

To present our definition, it is convenient to specify a partial order.

Assume that $S = (U, A)$ is an information system with $|U| = n$. Denote by $\bar{n}$ the set $\{1, 2, \ldots, n\}$. For any vector $\vec{X} = (X_1, X_2, \ldots, X_n) \in \mathscr{P}(U)^n$, we define $\beta(\vec{X})$ to be the vector $(r_1, r_2, \ldots, r_n) \in \bar{n}^n$ that satisfies $r_1 \geq r_2 \geq \cdots \geq r_n$ and $\{\!\{r_1, r_2, \ldots, r_n\}\!\} = \{\!\{|X_1|, |X_2|, \ldots, |X_n|\}\!\}$. In other words, the function of $\beta$ is to sort the cardinalities of the components of $\vec{X}$ in descending order. Clearly, this is well-defined and thus $\beta$ is a mapping from $\mathscr{P}(U)^n$ to $\bar{n}^n$. For instance, $\beta(U, U, \ldots, U) = (n, n, \ldots, n)$. Recall that the mapping $\alpha : \mathscr{P}(A) \longrightarrow \mathscr{P}(U)^n$ maps $P$ to $(S_P(x_1), S_P(x_2), \ldots, S_P(x_{|U|}))$. The mapping $\beta$, together with $\alpha$, gives a composition mapping $\beta\alpha : \mathscr{P}(A) \longrightarrow \bar{n}^n$. For convenience, we call $\beta\alpha(P)$ the character of $P$ (with respect to the information systems $S$).

We are now in the position to define an equivalence relation on $\mathscr{P}(A)$. For any $P, Q \in \mathscr{P}(A)$, we say that $P \sim Q$ if and only if $P$ and $Q$ have the same character. Clearly, the binary relation $\sim$ is reflexive, symmetric, and transitive. Therefore, $\sim$ is an equivalence relation and $\mathscr{P}(A)/\sim$ yields a partition of $\mathscr{P}(A)$. We write $[P]$ for the induced equivalence class containing $P$, that is, $[P] = \{Q \subseteq A \mid \beta\alpha(Q) = \beta\alpha(P)\}$.

Further, let us define a partial order $\leq$ on $\mathscr{P}(A)/\sim$: For any $[P], [Q] \in \mathscr{P}(A)/\sim$, $[P] \leq [Q]$ if and only if $\beta\alpha(P) \leq \beta\alpha(Q)$, where $\leq$ holds if and only if every component of $\beta\alpha(P)$ is not greater than the corresponding component of $\beta\alpha(Q)$. Formally, we write $(r_1, r_2, \ldots, r_n) \leq (s_1, s_2, \ldots, s_n)$ if and only if $r_i \leq s_i$ for every $i \in \{1, 2, \ldots, n\}$. We also write $(r_1, r_2, \ldots, r_n) < (s_1, s_2, \ldots, s_n)$ if and only if $(r_1, r_2, \ldots, r_n) \leq (s_1, s_2, \ldots, s_n)$ and $r_i < s_i$ for some $i \in \{1, 2, \ldots, n\}$. By definition, $\leq$ is indeed a partial order on $\mathscr{P}(A)/\sim$. If $[P] \leq [Q]$ and $[P] \neq [Q]$, then we say that $[P]$ is weaker than $[Q]$, denoted by $[P] < [Q]$, which is equivalent to that $\beta\alpha(P) < \beta\alpha(Q)$.

**Definition 5.** *Let $S = (U, A)$ be an information system and $G$ a function from $\mathscr{P}(A)$ to $\mathbb{R}$, the set of real numbers. We say that $G$ is an information granulation of $S$ if $G$ satisfies the following conditions:*

*(1) $G(P) \geq 0$ for any $P \in \mathscr{P}(A)$ (Nonnegativity).*
*(2) $G(P) = G(Q)$ for any $P, Q \in \mathscr{P}(A)$ with $[P] = [Q]$ (Invariability).*
*(3) $G(P) < G(Q)$ for any $P, Q \in \mathscr{P}(A)$ with $[P] \prec [Q]$ (Monotonicity).*

Like all the existing definitions of information granulation, the granulation value $G(P)$ requires nonnegative in the above definition. The second condition just says that if two subsets are of the same character, then they have the same granulation value. The third condition says that information granulation is strictly monotonic on characters.

**Remark 1.**

1) Compared with Definition 1, our definition avoids the use of mappings between multisets. Furthermore, the relation $\prec$ here is weaker than $\prec''$ in Definition 1, namely, $P \prec'' Q$ implies $[P] \prec [Q]$.
2) By comparison with Definition 2, our information granulation is impossible to take constant values. In addition, our definition avoids the use of mappings between multisets.
3) In contrast with Definition 3, the mapping $G$ here equates more values just because $\alpha(P) = \alpha(Q)$ implies $[P] = [Q]$, but the converse may not hold. For example, if $[P] = [Q]$ and $\alpha(P) \neq \alpha(Q)$, Definition 5 forces that $G(P) = G(Q)$, which cannot be obtained from Definition 3. Moreover, the order $\prec$ here is weaker than $\prec'$ in Definition 3.
4) Compared with Definition 4, Definition 5 is applicable to both complete and incomplete information systems. Notice that the mapping $G$ in Definition 4 has 0 and 1 as its minimum and maximum, respectively; this requirement is absent from Definition 5. In fact, it is not essential since we can transform any information granulation $G$ satisfying Definition 5 into a bounded one. Concretely, note that the set $\{G(P) \mid P \subseteq A\}$ is finite, so $G$ has the minimum, say, $m$, and the maximum, say, $M$. Since $G$ is not constant, we see that $M > m \geq 0$. Setting

$$G'(P) = \frac{G(P) - m}{M - m},$$

we find that $G'$ is an information granulation in the sense of Definition 5, and moreover, $G'$ can achieve the maximum 1 and the minimum 0.

*4.2. Case study*

It has been demonstrated that all the existing concrete information granulations [2, 6, 7, 9, 10] fulfil Definitions 1, 2, 3, and 4 well. As an example, in this section we directly show that two information granulations in [2, 6] are the instances of Definition 5. More instances will be discussed in the next section.

The following information granulation is defined for a complete information system.

**Definition 6** ([6])**.** *Let $S = (U, A)$ be a complete information system. Suppose that $P \subseteq A$ such that $U/Ind(P) = \{X_1, X_2, \ldots, X_m\}$. Then the information granulation of $P$ is defined as*

$$GK(P) = \frac{1}{|U|^2} \sum_{i=1}^{m} |X_i|^2.$$

As expected, we have the following observation.

**Theorem 1.** *GK in Definition 6 is an information granulation in the sense of Definition 5.*

PROOF. Let us check the conditions in Definition 5 one by one.
1) By the definition of $GK$, the first condition is satisfied.

2) Suppose that $P, Q \in \mathscr{P}(A)$ such that $[P] = [Q]$. It follows that $\{|S_P(x_1)|, |S_P(x_2)|, \ldots, |S_P(x_{|U|})|\} = \{|S_Q(x_1)|, |S_Q(x_2)|, \ldots, |S_Q(x_{|U|})|\}$. Let $U/Ind(P) = \{X_1, X_2, \ldots, X_m\}$ and $U/Ind(Q) = \{Y_1, Y_2, \ldots, Y_n\}$. By eqs. (1)-(3), we have that

$$\begin{aligned} GK(P) &= \frac{1}{|U|^2} \sum_{i=1}^{m} |X_i|^2 \\ &= \frac{1}{|U|^2} \sum_{i=1}^{m} \sum_{j=1}^{s_i} |S_P(x_{ij})| \\ &= \frac{1}{|U|^2} \sum_{x \in U} |S_P(x)| \\ &= \frac{1}{|U|^2} \sum_{x \in U} |S_Q(x)| \\ &= \frac{1}{|U|^2} \sum_{j=1}^{n} |Y_j|^2 \\ &= GK(Q), \end{aligned}$$

namely, $GK(P) = GK(Q)$, as desired.

3) Assume that $P, Q \in \mathscr{P}(A)$ such that $[P] = (p_1, p_2, \ldots, p_{|U|}) \prec [Q] = (q_1, q_2, \ldots, q_{|U|})$. Then there exists $k \in \{1, 2, \ldots, |U|\}$ such that $p_k < q_k$ and $p_j \le q_j$ for all $j \in \{1, 2, \ldots, |U|\}$ with $j \ne k$. Note that $\{p_1, p_2, \ldots, p_{|U|}\} = \{|S_P(x_1)|, |S_P(x_2)|, \ldots, |S_P(x_{|U|})|\}$ and $\{q_1, q_2, \ldots, q_{|U|}\} = \{|S_Q(x_1)|, |S_Q(x_2)|, \ldots, |S_Q(x_{|U|})|\}$. We thus get by eqs. (1)-(3) that

$$\begin{aligned} GK(P) &= \frac{1}{|U|^2} \sum_{x \in U} |S_P(x)| \\ &= \frac{1}{|U|^2} \sum_{i=1}^{|U|} p_i \\ &< \frac{1}{|U|^2} \sum_{i=1}^{|U|} q_i \\ &= \frac{1}{|U|^2} \sum_{x \in U} |S_Q(x)| \\ &= \frac{1}{|U|^2} \sum_{j=1}^{n} |Y_j|^2 \\ &= GK(Q), \end{aligned}$$

namely, $GK(P) < GK(Q)$. Therefore, $GK$ is an information granulation in the sense of Definition 5, finishing the proof of the theorem.

The next information granulation is defined for an incomplete information system.

**Definition 7** ([2]). *Let $S = (U, A)$ be an incomplete information system. Suppose that $P \subseteq A$ such that $\alpha(P) = (S_P(x_1), S_P(x_2), \ldots, S_P(x_{|U|}))$. Then the information granulation of $P$ is defined as*

$$GK(P) = \frac{1}{|U|^2} \sum_{i=1}^{|U|} |S_P(x_i)|.$$

Analogous to Theorem 1, we have the following result.

**Theorem 2.** *GK in Definition 7 is an information granulation in the sense of Definition 5.*

PROOF. Let us verify the conditions in Definition 5.

1) Clearly, the nonnegativity holds.

2) Suppose that $P, Q \in \mathscr{P}(A)$ such that $[P] = [Q]$. It follows that $\{|S_P(x_1)|, |S_P(x_2)|, \ldots, |S_P(x_{|U|})|\} = \{|S_Q(x_1)|, |S_Q(x_2)|, \ldots, |S_Q(x_{|U|})|\}$. Consequently, we see that

$$\begin{aligned} GK(P) &= \frac{1}{|U|^2} \sum_{i=1}^{|U|} |S_P(x_i)| \\ &= \frac{1}{|U|^2} \sum_{i=1}^{|U|} |S_Q(x_i)| \\ &= GK(Q), \end{aligned}$$

namely, $GK(P) = GK(Q)$. Hence, the invariability holds.

3) For the monotonicity, let $P, Q \in \mathscr{P}(A)$ such that $[P] \prec [Q]$, which is equivalent to that $\beta\alpha(P) < \beta\alpha(Q)$. Assume that $\beta\alpha(P) = (p_1, p_2, \ldots, p_{|U|})$ and $\beta\alpha(Q) = (q_1, q_2, \ldots, q_{|U|})$. Then there exists $k \in \{1, 2, \ldots, |U|\}$ such that $p_k < q_k$ and $p_j \leq q_j$ for all $j \in \{1, 2, \ldots, |U|\}$ with $j \neq k$. Note that $\{p_1, p_2, \ldots, p_{|U|}\} = \{|S_P(x_1)|, |S_P(x_2)|, \ldots, |S_P(x_{|U|})|\}$ and $\{q_1, q_2, \ldots, q_{|U|}\} = \{|S_Q(x_1)|, |S_Q(x_2)|, \ldots, |S_Q(x_{|U|})|\}$. We thus obtain that

$$\begin{aligned} GK(P) &= \frac{1}{|U|^2} \sum_{i=1}^{|U|} |S_P(x_i)| \\ &= \frac{1}{|U|^2} \sum_{i=1}^{|U|} p_i \\ &< \frac{1}{|U|^2} \sum_{i=1}^{|U|} q_i \\ &= \frac{1}{|U|^2} \sum_{i=1}^{|U|} |S_Q(x_i)| \\ &= GK(Q), \end{aligned}$$

namely, $GK(P) < GK(Q)$. Therefore, $GK$ is an information granulation in the sense of Definition 5, as desired.

**Remark 2.** We remark that Theorems 10, 11, 12, 14, and 15 in [5] still hold if the order $\prec'$ is relaxed to $\prec$. The proofs are not difficult, and thus we omit the details here.

## 5. A universal construction of information granulation

In this section, we establish a universal construction of information granulation. As a by-product of the construction, we see that each information granulation in [2, 6, 7, 9, 10] is an instance of our axiomatic definition introduced in the last section.

We first relate information granulations in the sense of Definition 5 with a class of functions of multiple variables, which will provide a universal construction of concrete information granulation.

Let $f : \bar{n}^n \longrightarrow \mathbb{R}$ be a real-valued function of $n$ variables, where $\bar{n} = \{1, 2, \ldots, n\}$. We say that $f$ is monotonic if $f$ is strictly monotonically increasing on each of its variables.

In the next theorem, we keep the definitions of mappings $\alpha : \mathscr{P}(A) \longrightarrow \mathscr{P}(U)^n$ and $\beta : \mathscr{P}(U)^n \longrightarrow \bar{n}^n$ given in Section 4.1.

**Theorem 3.** *Let $S = (U, A)$ be an information system with $|U| = n$.*

*(1) For any monotonic function $f : \bar{n}^n \longrightarrow \mathbb{R}$, there is an information granulation $G_f$ of $S$ such that $G_f - f\beta\alpha$ is constant.*

*(2) For any information granulation $G$ of $S$, there is a monotonic function $f_G : \bar{n}^n \longrightarrow \mathbb{R}$ such that $f_G \beta \alpha = G$.*

To prove the above theorem, it is convenient to have the following lemma.

**Lemma 1.** *Let $(X, \leq_0)$ be a finite, partially ordered set and $Y$ a subset of $X$ inheriting the ordering $\leq_0$. If the function $g : Y \longrightarrow \mathbb{R}$ is monotonic, then there exists an extension $\hat{g} : X \longrightarrow \mathbb{R}$ of $g$ satisfying that $\hat{g}$ is monotonic and $\hat{g}|_Y = g$, where the notation $\hat{g}|_Y = g$ represents the restriction of $\hat{g}$ to $Y$.*

PROOF. Suppose that $X = \{x_1, x_2, \ldots, x_m\}$. By the well-known order-extension principle, the partial order $\leq_0$ on $X$ can be extended to a total order, say, $\leq_0$, on $X$ such that $x \leq_0 y$ implies $x \leq_0 y$. Without loss of generality, we may assume that $x_1 \leq_0 x_2 \leq_0 \cdots \leq_0 x_m$. Let us define the extension $\hat{g}$ as follows. If $x_1 \in Y$, then set $\hat{g}(x_1) = g(x_1)$; otherwise, suppose that $i_1$ is the smallest index such that $x_{i_1} \in Y$, that is, $x_j \notin Y$ for any $1 \leq j < i_1$. Define $\hat{g}(x_{i_1}) = g(x_{i_1})$, and for each $1 \leq j < i_1$, set

$$\hat{g}(x_j) = g(x_{i_1}) - \frac{i_1 - j}{i_1}.$$

Further, suppose that $i_2$ is the second-smallest index such that $x_{i_2} \in Y$, that is, $x_j \notin Y$ for any $i_1 < j < i_2$. Then we set $\hat{g}(x_{i_2}) = g(x_{i_2})$, and for each $i_1 < j < i_2$, set

$$\hat{g}(x_j) = g(x_{i_2}) - (i_2 - j)\frac{g(x_{i_2}) - g(x_{i_1})}{i_2 - i_1}.$$

Let us proceed in this way. Finally, suppose that $i_s$ is the largest index such that $x_{i_s} \in Y$ and $\hat{g}(x_j)$ has been defined for all $j \leq i_s$. If $i_s = m$, then we can stop; otherwise, for $i_s < j \leq m$ set $\hat{g}(x_j) = g(x_{i_s}) + j$, which finishes the definition of $\hat{g}$. It is very easy to verify that the resultant $\hat{g}$ satisfies the two requirements of the lemma, finishing the proof.

With the lemma above, we can now prove Theorem 3.

**Proof of Theorem 3.** (1) Let $f : \bar{n}^n \longrightarrow \mathbb{R}$ be a monotonic function. Then the image of $f$ is finite since the domain of $f$ has only $n^n$ elements. Hence, $f$ has a minimum value, denoted by $\min(f)$. Let us take an arbitrary real number $c \geq -\min(f)$ and define

$$G_f(P) = f\beta\alpha(P) + c$$

for any $P \in \mathscr{P}(A)$. It remains only to check that $G_f$ satisfies the three conditions in Definition 5. Firstly, by definition we see that $G_f(P) \geq \min(f) + c \geq 0$, i.e., $G_f(P) \geq 0$ for any $P \subseteq A$, so the nonnegativity holds. Secondly, let $[P] = [Q]$. Then we have that $\beta\alpha(P) = \beta\alpha(Q)$, and thus $f\beta\alpha(P) = f\beta\alpha(Q)$, which means that $G_f(P) = G_f(Q)$. Thirdly, assume that $[P] \prec [Q]$, which is equivalent to that $\beta\alpha(P) < \beta\alpha(Q)$. We thus get by the monotonicity of $f$ that $f\beta\alpha(P) < f\beta\alpha(Q)$. It implies that $G_f(P) < G_f(Q)$, as desired.

(2) Let $G$ be an information granulation of $S$. We now define $f_G$ by three steps.
Step 1. Define $g : Im(\beta\alpha) \longrightarrow \mathbb{R}$ as follows. Let $\vec{r} = (r_1, r_2, \ldots, r_n) \in Im(\beta\alpha)$. Then there is $P_{\vec{r}} \in \mathscr{P}(A)$ such that $\beta\alpha(P_{\vec{r}}) = \vec{r}$. We set $g(\vec{r}) = G(P_{\vec{r}})$. This is well-defined because for any $Q \in \mathscr{P}(A)$ with $\beta\alpha(Q) = \vec{r}$, we always have that $G(Q) = G(P_{\vec{r}})$ by the invariability of $G$. It follows from the monotonicity of $G$ that $g$ is monotonic with respect to the partial order $\leq$ defined in Section 4.1.
Step 2. Extend $g$ to $\hat{g} : \mathcal{N} \longrightarrow \mathbb{R}$, where $\mathcal{N} = \{(r_1, r_2, \ldots, r_n) \in \bar{n}^n \mid r_1 \geq r_2 \geq \cdots \geq r_n\}$ endowed with the partial order $\leq$. This follows immediately by Lemma 1.
Step 3. Define $f_G : \bar{n}^n \longrightarrow \mathbb{R}$ as follows. For any $\vec{r} = (r_1, r_2, \ldots, r_n) \in \bar{n}^n$, let $\pi(\vec{r}) = (s_1, s_2, \ldots, s_n)$, in which $s_1 \geq s_2 \geq \cdots \geq s_n$ and $\{\!\{s_1, s_2, \ldots, s_n\}\!\} = \{\!\{r_1, r_2, \ldots, r_n\}\!\}$. Further, set $f_G(\vec{r}) = \hat{g}(\pi(\vec{r}))$.

It is quite straightforward to show that the resultant function $f_G$ is monotonic and $f_G \beta\alpha = G$. This completes the proof of the theorem.

Clearly, the assertion (1) in Theorem 3 provides us an approach to the construction of concrete information granulation. Let us see several examples.

**Example 1.** Let $S = (U, A)$ be an information system with $U = \{x_1, x_2, \ldots, x_n\}$. Observe that $f : \bar{n}^n \longrightarrow \mathbb{R}$ defined by

$$f(r_1, r_2, \ldots, r_n) = \frac{1}{n^2}\sum_{i=1}^{n} r_i$$

is a monotonic function. Obviously, $\min(f) = 1/n$. For any $P \in \mathscr{P}(A)$, by definition we always have that

$$f\beta\alpha(P) = \frac{1}{n^2} \sum_{i=1}^{n} |S_P(x_i)|.$$

It follows from Theorem 3 that for any $c \geq -1/n$, defining

$$G_f(P) = f\beta\alpha(P) + c = \frac{1}{n^2} \sum_{i=1}^{n} |S_P(x_i)| + c$$

gives rise to an information granulation of $S$ in the sense of Definition 5. In particular, taking $c = 0$ we see that

$$G_f(P) = \frac{1}{n^2} \sum_{i=1}^{n} |S_P(x_i)|,$$

which is exactly the information granulation defined in [2]. When $S$ is complete, the information granulation reduces to the one in [6]. This means that both of the information granulations in [2, 6] satisfy our axiomatic definition.

**Example 2.** Suppose that $S = (U, A)$ is an information system with $U = \{x_1, x_2, \ldots, x_n\}$. Define $f : \bar{n}^n \longrightarrow \mathbb{R}$ by

$$f(r_1, r_2, \ldots, r_n) = \frac{1}{n} \sum_{i=1}^{n} \frac{r_i(r_i - 1)}{n(n-1)}.$$

Then it is clear that $f$ is a monotonic function and $\min(f) = 0$. For any $P \in \mathscr{P}(A)$, by definition we always have that

$$f\beta\alpha(P) = \frac{1}{n} \sum_{i=1}^{n} \frac{|S_P(x_i)|(|S_P(x_i)| - 1)}{n(n-1)}.$$

It follows from Theorem 3 that for any $c \geq 0$, defining

$$\begin{aligned} G_f(P) &= f\beta\alpha(P) + c \\ &= \frac{1}{n} \sum_{i=1}^{n} \frac{|S_P(x_i)|(|S_P(x_i)| - 1)}{n(n-1)} + c \end{aligned}$$

yields an information granulation of $S$ in the sense of Definition 5. In particular, taking $c = 0$ we get that

$$G_f(P) = \frac{1}{n} \sum_{i=1}^{n} \frac{|S_P(x_i)|(|S_P(x_i)| - 1)}{n(n-1)},$$

which is exactly the combination granulation defined in [9]. When the information system $S$ is complete, the information granulation degenerates to the combination granulation in [10]. Therefore, the information granulations defined in [9, 10] fulfil our axiomatic definition.

**Example 3.** Let $S = (U, A)$ be an information system with $U = \{x_1, x_2, \ldots, x_n\}$. We now define $f : \bar{n}^n \longrightarrow \mathbb{R}$ by

$$f(r_1, r_2, \ldots, r_n) = \frac{1}{n} \sum_{i=1}^{n} \log_2 r_i.$$

Then it is evident that $f$ is a monotonic function and $\min(f) = 0$. For any $P \in \mathscr{P}(A)$, by definition we always have that

$$f\beta\alpha(P) = \frac{1}{n} \sum_{i=1}^{n} \log_2 |S_P(x_i)|.$$

It follows from Theorem 3 that for any $c \geq 0$, defining

$$\begin{aligned} G_f(P) &= f\beta\alpha(P) + c \\ &= \frac{1}{n} \sum_{i=1}^{n} \log_2 |S_P(x_i)| + c \end{aligned}$$

yields an information granulation of $S$ in the sense of Definition 5. In particular, taking $c = 0$ we get that

$$G_f(P) = \frac{1}{n} \sum_{i=1}^{n} \log_2 |S_P(x_i)|,$$

which is exactly the rough entropy defined in [7]. When the information system $S$ is complete, the information granulation degenerates to the rough entropy in [2, 6]. In other words, each information granulation defined in [2, 6, 7] satisfies our axiomatic definition.

## 6. Conclusion

In this paper, we have developed an improved axiomatic definition of information granulations and compared it with the ones in the literature. Furthermore, we provided a universal construction of concrete information granulation by making a connection between information granulations and a class of functions of multiple variables. As expected, we have proven that our axiomatic definition has some concrete information granulations in the literature as instances.

## Acknowledgements

This work was supported by the National Natural Science Foundation of China.